\documentclass[11pt,a4paper]{article}
\usepackage[hyperref]{emnlp2020}
\usepackage{times}
\usepackage{latexsym}

\usepackage{booktabs}
\usepackage{xspace}
\usepackage{amsmath,amssymb}
\usepackage{xcolor}
\usepackage{multirow}
\usepackage{rotating}
\usepackage{paralist}
\usepackage{tabularx}
\usepackage{balance}
\usepackage[utf8]{inputenc}
\usepackage{microtype}

\DeclareUnicodeCharacter{2212}{−}

\newcommand{\parc}{\textsc{parc}\xspace}
\newcommand{\onto}{OntoNotes\xspace}
\newcommand{\riqua}{\textsc{RiQuA}\xspace}
\newcommand{\conll}{CoNLL'00\xspace}
\newcommand{\chem}{ChemDNer\xspace}
\newcommand{\nspans}{[freq]\xspace}
\newcommand{\spanlen}{[span length]\xspace}

\newcommand{\bounddiv}{[boundary distinctness]\xspace}
\newcommand{\fone}{$F_1$\xspace}

\aclfinalcopy
\def\aclpaperid{2438}

\title{Dissecting Span Identification Tasks with Performance Prediction}

\author{Sean Papay, Roman Klinger, \and Sebastian Pad\'o\\
    Institut f{\"u}r Maschinelle Sprachverarbeitung, University of Stuttgart \\
    Pfaffenwaldring 5b, 70569 Stuttgart, Germany \\
    \texttt{\{sean.papay,klinger,pado\}@ims.uni-stuttgart.de}
    }

\date{}

\begin{document}
\maketitle

\begin{abstract}
  Span identification (in short, span ID) tasks such as chunking, NER,
  or code-switching detection, ask models to identify and classify relevant
  spans in a text. Despite being a staple of NLP, and sharing a common
  structure, there is little insight on how these tasks' properties influence
  their difficulty, and thus little guidance on what model families
  work well on span ID tasks, and why.  We analyze span ID tasks via
  \textit{performance prediction}, estimating how well neural
  architectures do on different tasks.

  Our contributions are: (a) we identify key properties of span ID
  tasks that can inform performance prediction; (b) we carry out a
  large-scale experiment on English data, building a model to predict
  performance for unseen span ID tasks that can support architecture
  choices; (c), we investigate the parameters of the meta model,
  yielding new insights on how model and task properties interact to
  affect span ID performance. We find, e.g., that span frequency is
  especially important for LSTMs, and that CRFs help when spans are
  infrequent and boundaries non-distinctive.
\end{abstract}

\section{Introduction}
\textit{Span identification} is a family of analysis tasks that make
up a substantial portion of applied NLP. Span identification (or short,
span ID) tasks have in common that they identify and classify
contiguous spans of tokens within a running text. Examples are named
entity recognition \citep{nadeau2007survey}, chunking
\citep{sang2000introduction}, entity extraction
\citep{etzioni2005unsupervised}, quotation
detection~\citep{pareti2016parc}, keyphrase detection
\citep{augenstein-etal-2017-semeval}, or code switching
\citep{pratapa-etal-2018-language}.
In terms of complexity, span ID tasks form a middle ground between
simpler analysis tasks that predict labels for single linguistic units
(such as lemmatization \cite{porter1980algorithm} or sentiment
polarity classification \cite{Liu2012}) and more complex analysis
tasks such as relation extraction, which combines span ID with
relation identification
\citep{zelenko-etal-2002-kernel,adel2018domainindependent}.

Due to the rapid development of deep learning, an abundance of model
architectures is available for the implementation of span ID tasks.
These include isolated token classification models
\cite{berger-etal-1996-maximum,chieu-ng-2003-named}, probabilistic
models such as hidden Markov models \cite{Rabiner1989}, maximum
entropy Markov models \cite{mccallum2000maximum}, and conditional
random fields \cite{Lafferty2001}, recurrent neural networks such as
LSTMs \cite{hochreiter1997long}, and transformers such as BERT
\cite{devlin19:_bert}.

Though we have some understanding what each of these models can and
cannot learn, there is, to our knowledge, little work on
systematically understanding how different span ID tasks compare: are
there model architectures that work well generally? Can we identify
properties of span ID tasks that can help us select suitable model
architectures on a task-by-task basis? Answers to these questions
could narrow the scope of architecture search for these tasks, and
could help with comparisons between existing methods and more recent
developments.

In this work, we address these questions by applying
\textit{meta-learning} to span identification
\citep{vilalta2002perspective, vanschoren2018meta}. Meta-learning
means ``systematically observing how different machine learning
approaches perform [\dots] to learn new tasks much faster''
\citep{vanschoren2018meta}, with examples such as architecture search
\citep{elsken2018neural} and hyperparameter optimization
\citep{bergstra2012random}. Our specific approach is to apply
\textit{performance prediction} for span ID tasks, using both task
properties and model architectures as features, in order to obtain a
better understanding of the differences among span ID tasks.

Concretely, we collect a set of English span ID tasks, quantify key
properties of the tasks (such as how distinct the spans are from their
context, and how clearly their boundaries are marked) and formulate
hypotheses linking properties to performance
(Section~\ref{sec:tasks}). Next, we describe relevant neural model
architectures for span ID (Section~\ref{sec:architectures}). We then
train a linear regressor as a meta-model to predict span ID
performance based on model features and task metrics in an unseen-task
setting (Section~\ref{sec:meta-learning-model}). We find the best of these
architectures perform at or close to the state of the art, and their
success can be relatively well predicted by the meta-model
(Section~\ref{sec:experiments}). Finally, we carry out a detailed
analysis of the regression model's parameters
(Section~\ref{sec:analysis}), gaining insight into the relationship
between span ID tasks and different neural model architectures. For
example, we establish that spans that are not very distinct from their
context are consistently difficult to identify, but that CRFs are
specifically helpful for this class of span ID tasks.

\section{Datasets, Span Types, and Hypotheses}
\label{sec:tasks}

We work with five widely used English span ID datasets. All of them
have non-overlapping spans from a closed set of span types.  In the
following, we discuss (properties of) span types, assuming that each
span type maps onto one span ID task.

\subsection{Datasets}
\paragraph{Quotation Detection: \parc and \riqua.}
The Penn Attribution Relation Corpus (\parc) version 3.0 
\citep{pareti2016parc} and the Rich Quotation Attribution Corpus
\citep[\riqua,][]{papay-pad:2020:LREC} are two datasets for quotation
detection: models must identify direct and indirect quotation spans in
text, which can be useful for social network construction
\cite{Elson2010} and coreference resolution \cite{E14-1005}. The
corpora cover articles from the Penn Treebank (\parc) and 19th century English
novels (\riqua), respectively. Within each text, quotations are
identified, along with each quotation's speaker (or source), and its
cue (an introducing word, usually a verb like ``said'').  We model
detection of quotations as well as cues.  As speaker and addressee
identification are relation extraction tasks, we exclude these span
types.

\paragraph{Chunking: \conll.}
Chunking (shallow parsing) is an important preprocessing step in
a number of NLP applications. We use the corpus from the 2000 CoNLL
shared task on chunking (\conll) \citep{sang2000introduction}. Like
\parc, this corpus consists of a subset of the PTB.  This dataset is
labeled with non-overlapping chunks of eleven phrase types. In our
study, we consider the seven phrase types with $>$100 instances in the
training partition: `ADJP', `ADVP', `NP', `PP', `PRT', `SBAR', and `VP'.
\paragraph{NER: \onto and \chem.}
For recognition and classification of proper names, we use the NER
layer of OntoNotes Corpus v5.0 \citep{weischedel2013ontonotes} and
Biocreative's \chem corpus v1.0
\citep{krallinger2015chemdner}. OntoNotes, a general language NER
corpus, is our largest dataset, with over 2.2 million tokens. The NER
layer comprises 18 span types, both typical entity types such as
`Person' and `Organization' as well as numerical value types such as
`Date' and `Quantity'. We use all span types. \chem is a NER corpus
specific to chemical and drug names, comprising titles and abstracts
from 10000 PubMed articles. It labels names of chemicals and drugs and
assigns them to eight classes, corresponding to chemical
name nomenclatures. We use seven span types: `Abbreviation', `Family',
`Formula', `Identifier', `Systematic', `Trivial', and `Multiple'. We
exclude the class `No class' as infrequent ($<$100 instances).

\subsection{Span Type Properties and Hypotheses}

While quotation detection, chunking, and named entity recognition
are all span ID tasks, they vary quite widely in
their properties. As mentioned in the introduction, we know of
little work on quantifying the similarities and differences of span types,
and thus, span ID tasks.

\begin{table}
\centering
\setlength\tabcolsep{3.5pt}
\begin{tabular}{llrrrr}
\toprule
Task & Dataset & freq & len & SD & BD\\
\cmidrule(r){1-1}\cmidrule(r){2-2}\cmidrule(r){3-3}\cmidrule(r){4-4}\cmidrule(r){5-5}\cmidrule{6-6}
Quotation & \parc & 16480 & 7.89 & 1.34 & 1.43 \\
Quotation & \riqua & 4026 & 9.84 & 1.46 & 1.57 \\
Chunking & \conll & 37168 & 1.55 & 1.26 & 0.64 \\
NER & \chem & 6110 & 1.62 & 3.08 & 0.96 \\
NER & \onto & 16861 & 1.63 & 3.36 & 1.00 \\
\bottomrule
\end{tabular}

\caption{Span type metrics (values averaged over all span
  types in each corpus, weighted by span type frequency, computed on
  training sets). SD = span distinctiveness, BD = boundary
  distinctiveness. Values for individual span types can be found in Table
  \ref{tab:full_tasks} in the Appendix.}
\label{tab:tasks}
\end{table}

We now present four metrics which we propose to capture the relevant
characteristics of span types, and make concrete our hypotheses
regarding their effect on model performance. Table \ref{tab:tasks}
reports frequency-weighted averages for each metric on each
dataset. See Table~\ref{tab:full_tasks} in the Appendix for
all span-type-specific values.

\textbf{Frequency} is the number of spans for a span type in the
dataset's training corpus.  It is well established that the
performance of a machine learning model benefits from higher amounts
of training data \citep{halevy2009unreasonable}.  Thus, we expect this
property to be positively correlated with performance.  However, some
architectural choices, such as the use of transfer learning, are
purported to reduce the data requirements of machine learning models
\citep{pan2009survey}, so we expect a smaller correlation for
architectures which incorporate transfer learning.

\textbf{Span length} is the geometric mean of spans' lengths, in
tokens.  \newcite{scheible2016model} note that traditional CRF models
perform poorly at the identification of long spans due to the strict
Markov assumption they make \cite{Lafferty2001}. Thus, we expect
architectures which rely on such assumptions and which have no way to
model long distance dependencies to perform poorly on span types with
a high average span length, while LSTMs or transformers should do
better on long spans \citep{khandelwal2018sharp,
  vaswani2017attention}.

\textbf{Span distinctiveness} is a measure
of how distinctive the text that comprises spans is compared to the
overall text of the corpus.  Formally, we define it as the KL
divergence $D_{\text{KL}}(P_{\text{span}}||P)$, where $P$ is the
unigram word distribution of the corpus, and $P_{\text{span}}$ is
the unigram distribution of tokens within a span.  A high span
distinctiveness indicates that different words are used inside spans
compared to the rest of the text, while a low span distinctiveness
indicates that the word distribution is similar inside and outside of
spans.
\par We expect this property to be positively correlated with model
performance.  Furthermore, we hypothesize that span types with a
high span distinctiveness should be able to rely more heavily on
local features, as each token carries strong information about span
membership, while low span distinctiveness calls for sequence information.
Consequently, we expect that architectures incorporating
sequence models such as CRFs, LSTMs, and transformers should perform
better at low-distinctive span types.
   
\textbf{Boundary distinctiveness} is a measure of how distinctive the
starts and ends of spans are.  We formalize this in terms of a
KL-divergence as well, namely as
$D_{\text{KL}}(P_{\text{boundary}}||P)$ between the unigram word
distribution ($P$) and the distribution of boundary tokens ($P_{\text{boundary}}$),
where boundary tokens are those which occur immediately before the start of a span, or
immediately after the end of a span.  A high
boundary distinctiveness indicates that the start and end points of
spans are easy to spot, while low distinctiveness indicates smooth
transitions.
\par We expect boundary distinctiveness to be positively correlated with
model performance, based on studies that obtained improvements from
specifically modeling the transition between span and context
\cite{todorovic2008named,scheible2016model}. As
sequence information is required to utilize boundary information,
high boundary distinctiveness should improve performance more for
LSTMs, CRFs, or transformers.

\paragraph{Task Profiles.} As Table~\ref{tab:tasks} shows, the
metrics we propose appear to capture the task structure of the
datasets well: quotation corpora have long spans with low span
distinctiveness (anything can be said) but high boundary
distinctiveness (punctuation, cues). Chunking has notably low
boundary distinctiveness, due to the syntactic nature of the span
types, and NER spans show high distinctiveness (semantic classes)
but are short and have somewhat indistinct boundaries as well.

\section{Model Architectures}
\label{sec:architectures}
For span identification, we use the BIO framework
\citep{ramshaw1999text}, framing span identification as a sequence
labeling task. As each span type has its own B and I labels, and there
is one O label, a dataset with $n$ span types leads to a $2n+1$-label
classification problem for each token.

We investigate a set of sequence labeling models, ranging from
baselines to state-of-the-art architectures. We group our models by
common components, and build complex models through combination of
simpler models. Except for the models using BERT, all architectures assume one
300-dimensional GloVe embedding \citep{pennington2014glove} per token
as input.

\paragraph{Baseline.}
As a baseline model, we use a simple token-level classifier.  This
architecture labels each token using a softmax classifier without
access to sequence information (neither at the label level nor at
the feature level).

\paragraph{CRF.} This model uses a linear-chain conditional random
field (CRF) to predict token label sequences \citep{Lafferty2001}. It
can access neighboring labels in the sequence of predictions.

\paragraph{LSTM and LSTM+CRF.}
These architectures incorporate Bi-directional LSTMs (biLSTMs)
\citep{hochreiter1997long, schuster1997bidirectional} as components.
The simplest architecture, LSTM, passes the inputs through a 2-layer
biLSTM network, and then predicts token labels using a softmax layer.
The LSTM+CRF architecture combines the biLSTM network with a CRF
layer, training all weights simultaneously. These models can learn to
combine sequential input and labeling information.

\paragraph{BERT and BERT+CRF.}
These architectures include the pre-trained BERT language model
\citep{devlin19:_bert} as a component.  The simplest architecture in
this category, BERT, comprises a pre-trained BERT encoder and a
softmax output layer, which is trained while the BERT encoder is
fine-tuned.  BERT+CRF combines a BERT encoder with a linear-chain CRF
output layer, which directly uses BERT's output embeddings as inputs.
In this architecture, the CRF layer is first trained to convergence
while BERT's weights are held constant, and then both models are
jointly fine-tuned to convergence.  As BERT uses WordPiece
tokenization \citep{wordpiece}, the input must be re-tokenized
for BERT architectures.

\paragraph{BERT+LSTM+CRF.}
This architecture combines all components previously mentioned.  It
first uses a pre-trained BERT encoder to generate a sequence of
contextualized embeddings.  These embeddings are projected to 300
dimensions using a linear layer, yielding a sequence of vectors, which
are then used as input for a LSTM+CRF network. As with BERT+CRF, we first
train the non-BERT parameters to convergence while holding BERT's parameters
fixed, and subsequently fine-tune all parameters jointly.

\paragraph{Handcrafted Features.}
Some studies have shown marked increases in performance by adding
hand-crafted features \citep[e.g.][]{shimaoka-etal-2017-neural}. We
develop such features for our tasks and treat these to be an
additional architecture component. For architectures with this
component, a bag of features is extracted for each token (the exact
features used for each dataset are enumerated in Table\
\ref{tab:features} in the Appendix).  For each feature, we learn a
300-dimensional feature embedding which is averaged with the
GloVe or BERT embedding to obtain a token embedding. Handcrafted features can
be used with the Baseline, LSTM, LSTM+CRF, and BERT+LSTM+CRF
architectures. BERT and BERT+CRF cannot utilize manual features, as
they have no way of accepting token embeddings as input.

\section{Meta-learning Model}
\label{sec:meta-learning-model}

Recall that our meta-learning model is a model for predicting the
performance of the model architectures from Section
\ref{sec:architectures} when applied to span identification tasks from
Section \ref{sec:tasks}. We model this task of performance prediction
as linear regression, a well established framework for the statistical
analysis of language data \citep{baayen_2008}. The predictors are task
properties, model architecture properties, and their interactions, and
the dependent variable is (scaled) $F_1$ score.

While a linear model is not powerful enough to capture the full range
of interactions, its weights are immediately interpretable, it can
be trained on limited amounts of data, and it does not overfit easily
(see Section~\ref{sec:exper-proc}). All three properties make it a
reasonable choice for meta-learning.

\paragraph{Predictors and Interactions.}
As predictors for our performance prediction task, we use the span
type properties described above, and a number of binary model
properties.  For the span type properties \nspans and \spanlen, we use
the logarithms of these values as predictors.  The two distinctiveness
properties are already logarithms, and so we used them as-is.  For
model properties, we used four binary predicates: The presence of
handcrafted features, of a CRF output layer, of a bi-LSTM layer, and
of a BERT layer. 

In addition to main effects of properties of models and corpora on
performance (does a CRF layer help?), we are also interested in
interactions of these properties (does a CRF layer help in particular
for longer spans?). As such interactions are not captured
automatically in a linear regression model, we encode them as
predictors. We include interactions between span type and model
properties, as well as among model properties.

All predictors (including interactions) are
standardized so as to have a mean of zero and standard deviation of
one.

\paragraph{Scaling the Predicted Performance}
\label{sec:scal-pred-perf}
Instead of directly predicting the $F_1$ score, we instead make our
predictions in a logarithmic space, which eases the linearity
requirements of linear regression. We cannot directly use the logit
function to transform $F_1$ scores into
$F'=\text{logit}\left(\frac{F_1}{100}\right)$
since the presence of zeros in our $F_1$ scores makes this process
ill-defined.  Instead, we opted for a ``padded'' logit transformation
$F' = \text{logit}\left((1-\alpha)\cdot\frac{F_1}{100} +
  \alpha\cdot\frac{100-F_1}{100}\right)$ with a hyperparameter
$\alpha \in \left[0, 0.5\right)$.  This rescales the argument of the logit
function from $[0,1]$ to the smaller interval $[\alpha, 1-\alpha]$,
avoiding the zero problem of a
bare logit.  Through cross-validation
(cf. Section~\ref{sec:exper-proc}), we set $\alpha=0.2$.  We use the
inverse of this transformation to scale the output of our prediction
as an $F_1$ score, clamping the result to $[0, 100]$.

\section{Experiment}
\label{sec:experiments}
\subsection{Experimental Procedure}
\label{sec:exper-proc}

\begin{table*}[t]
\centering
\setlength\tabcolsep{3.5pt}
\begin{tabular}{lrrrrrrrrrrrr}
  \toprule
& BL & \shortstack[r]{Feat\\BL} & CRF & \shortstack[r]{Feat\\CRF} & LSTM & \shortstack[r]{Feat\\LSTM} & \shortstack[r]{LSTM\\CRF} & \shortstack[r]{Feat\\LSTM\\CRF} & BERT & \shortstack[r]{BERT\\CRF} & \shortstack[r]{BERT\\LSTM\\CRF} & \shortstack[r]{BERT\\Feat\\LSTM\\CRF}\\
\cmidrule(r){2-2}\cmidrule(r){3-3}\cmidrule(r){4-4}\cmidrule(r){5-5}\cmidrule(r){6-6}\cmidrule(r){7-7}\cmidrule(r){8-8}\cmidrule(r){9-9}\cmidrule(r){10-10}\cmidrule(r){11-11}\cmidrule(r){12-12}\cmidrule(r){13-13}
\parc & 34.8 & 31.4 & 46.6 & 57.9 & 64.8 & 76.9 & 76.0 & 81.8 & 78.7 & 81.4 & 82.4 & \textbf{82.5} \\
\riqua & 21.7 & 14.5 & 19.7 & 14.4 & 67.5 & 76.7 & 79.6 & 81.7 & 79.8 & \textbf{82.5} & 82.1 & 82.3 \\
\conll & 55.9 & 60.0 & 79.7 & 87.1 & 87.7 & 92.3 & 90.0 & 93.5 & 96.3 & 96.5 & 96.6 & \textbf{96.6} \\
\onto & 39.0 & 27.4 & 61.2 & 67.7 & 58.8 & 65.1 & 76.4 & 84.7 & 85.9 & 86.8 & 86.5 & \textbf{87.5} \\
\chem & 49.8 & 19.6 & 56.9 & 58.5 & 57.0 & 45.7 & 71.2 & 75.1 & 83.3 & 84.7 & \textbf{84.9} & 84.8 \\
\bottomrule
\end{tabular}

\caption{Average architecture results on datasets.  BL=Baseline,
  Feat=Hand-crafted features.  For each dataset, we micro-average
  performance over all span types, and average these micro-averages
  across five trials. For comparability with existing work, we include
  all span types in these micro-averages, even those which we exclude
  from our performance prediction.  Full performance results for each
  span type can be found in Table \ref{tab:full_results}.}
\label{tab:results}
\end{table*}

Our meta learning experiment comprises two steps: Span ID model
training, and meta model training.

\paragraph{Step 1: Span ID model training.} We train and subsequently
evaluate each model architecture on each dataset five times, using
different random initializations. With 12 model architectures and 5
datasets under consideration, this procedure yields
$12 \times 5 \times 5 = 300$ individual experiments.

For each dataset, we use the established train/test partition. Since
\riqua does not come with such a partition, we use cross-validation,
partitioning the dataset by its six authors and holding out one author
per cross-validation step.

We use early stopping for regularization, stopping training once
(micro-averaged) performance on a validation set reaches its maximum.
To prevent overfitting, all models utilize feature dropout -- during
training, each feature in a token's bag of input features is dropped
with a probability of 50\%. At evaluation time, all features are used.

\paragraph{Step 2: Meta learning model training.}
This step involves training our performance prediction model on the
$F_1$ scores obtained from the first step. For each
architecture-span-type pair of the 12 model architectures and 36 span
types, we already obtained 5 $F_1$ scores. This yields a total of
$12 \times 36 \times 5 = 2160$ input-output pairs to train our
performance prediction model.

We investigate both $L^1$ and $L^2$ regularization in an elastic net
setting \citep{zou2005regularization} but consistently find best
cross-validation performance with no regularization whatsoever.  Thus,
we use ordinary least squares regression.

To ensure that our performance predictions generalize, we use a
cross-validation setup when generating model predictions. To generate
performance predictions for a particular span type, we train our meta-model
on data from all other span types, holding out the span type for which we
want a prediction. We repeat this for all 36 span types, holding out a
different span type each time, in order to collect performance predictions
for each span type.

\subsection{Span Identification Results}

Step~1 yields 5 evaluation $F_1$ scores for each architecture--span-type
pair. This section summarizes the main findings. Detailed average
scores for each pair are reported in Table~\ref{tab:full_results} in
the Appendix.

Table \ref{tab:results} lists the micro-averaged performance of each
model architecture on each dataset.  Unsurprisingly,
BERT+Feat+LSTM+CRF, the model with the most components, performs best
on three of the five datasets.  This provides strong evidence that
this architecture can perform well across many tasks.  However, note
that architecture's dominance is somewhat overstated by only looking
at average dataset results. Our analysis permits us to look more
closely at results for individual span types, where we find that
BERT+Feat+LSTM+CRF performs best on 16 of the 36 total span types,
BERT+CRF on 7 span types, Feat+LSTM+CRF on 7 span types, and
BERT+LSTM+CRF on 6 span types. Thus, `bespoke' modeling of span types
can evidently improve results.

Even though our architectures are task-agnostic, and not tuned to
particular tasks or datasets, our best architectures still perform
quite competitively.  For instance, on \conll, our BERT+Feat+LSTM+CRF
model comes within 0.12 $F_1$ points of the best published model's
$F_1$ score of 97.62 \citep{akbik-etal-2018-contextual}.  For \parc,
existing literature does not report micro-averaged $F_1$ scores, but
instead focuses only on $F_1$ scores for content span detection.  In
this case, we find that our BERT+Feat+LSTM+CRF model beats the
existing state of the art on this span type, achieving an $F_1$ score
of 78.1, compared to the score of 75 reported in
\newcite{scheible2016model}.

\subsection{Meta-learning Results}

\begin{table}[t]
  \centering
\begin{tabularx}{\linewidth}{Xrr}
  \toprule
  & MAE & $r^2$\\
  \cmidrule(r){1-1}\cmidrule(rl){2-2}\cmidrule(l){3-3}
  Full model  & 11.38 & 0.73 \\
  No interactions & 14.00 & 0.61 \\
  Only architecture predictors  & 18.88 & 0.37  \\
  Only  task predictors & 20.87 & 0.22  \\                                 
  Empty model & 23.78  &  N/A \\
  \bottomrule
\end{tabularx}
\caption{Evaluation of performance prediction models}
\label{tab:errors}
\end{table}

\begin{figure}[t]
\centering
\includegraphics[width=\linewidth]{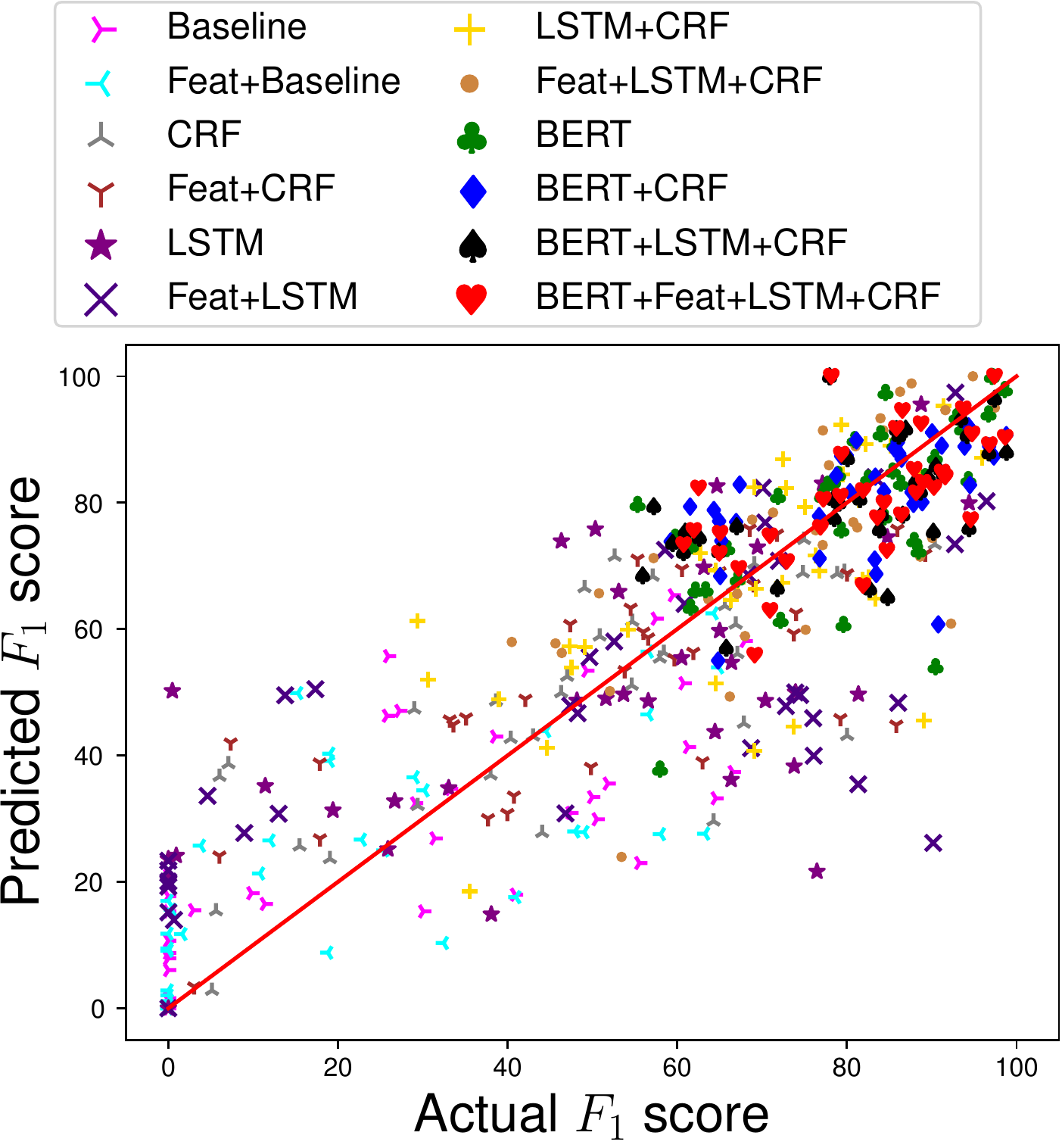}
\caption{Scatterplot of actual vs. predicted $F_1$
  scores for all 36 span types $\times$ 12 model architectures}
\label{fig:plot}
\end{figure}

The result of Step~2 is our performance prediction
model. Table~\ref{tab:errors} shows both mean absolute error (MAE),
which is directly interpretable as the mean difference between
predicted and actual F$_1$ score for each data point, and $r^2$, which
provides the amount of variance accounted for by the model. The full
performance prediction model, including both span type and model
architecture features, accounts for 73\% of the variance, with an MAE
of about 11. We see this as an acceptable model fit.  To validate the
usefulness of the predictor groups and interaction terms, we carry out
ablation experiments wherein these are excluded, including a model with
no interaction terms, a model with only span type-predictors, a model
with only architecture predictors, and an empty model, which only
predicts the average of all F$_1$ scores. The reduced models do better
than the empty model,\footnote{For the empty model, $r^2$ is undefined
  because the variance of the predictions is zero.} but show marked
increases in MAE and corresponding drops in $r^2$ compared to the full
model.  While the usefulness of the architecture predictors is
expected, this also constitutes strong evidence for the usefulness of
the span type predictors we have proposed in Section~\ref{sec:tasks}.

Figure \ref{fig:plot} shows a scatterplot of predicted and actual
$F_1$ scores.  Our meta learning model generally predicts high
performances better than low performances. The largest cluster of
errors occurs for experiments with an actual $F_1$ score of exactly
zero, arguably an uninteresting case. Thus, we believe that the
overall MAE underestimates rather than overestimates the quality of
the performance prediction for practical purposes.

\section{Analysis}
\label{sec:analysis}
We now investigate the linear regression coefficients of our
performance prediction model to assess our hypotheses from
Section~\ref{sec:tasks}. To obtain a single model to analyze, we
retrain our regression model on all data points, with no
cross-validation.

Table\ \ref{tab:coefs} shows the resulting coefficients. Using
Bonferroni correction at $\alpha=0.05$, we consider a coefficient significant if
p$<$0.002. Non-significant coefficients are shown in parentheses.  Due
to the scaling of $F_1$ scores performed as described in section
\ref{sec:scal-pred-perf}, the coefficients cannot be directly
interpreted in terms of linear change on the $F_1$ scale. However, as
we standardized all predictors, we can compare coefficients with one
another. Coefficients with a greater magnitude have larger effects on
$F_1$ score, with positive values indicating an increase, and negative
values a decrease.

When analyzing these coefficients, one must consider main effects and
interactions together. E.g.,\ the main effect coefficient for LSTMs is
negative, which seems to imply that adding an LSTM will hurt
performance. However, the LSTM $\times$ \nspans and LSTM $\times$
\bounddiv interactions are both strongly positive, so LSTMs should
help on frequent span types with high boundary distinctiveness. Our
main observations are the following:
\newcommand{\phcl}{\phantom{)}}
\begin{table}[tb!]
\centering
\begin{tabular}{llr}
\toprule
\multicolumn{3}{c}{\textbf{Model predictors}}\\
\midrule
Handcrafted & & ($-$0.11)\\
CRF & & 0.50\phcl\\
LSTM & & $-$0.35\phcl\\
BERT & & 1.00\phcl\\
\bottomrule
\multicolumn{3}{c}{\textbf{Span type predictors}}\\
\midrule
freq & & 0.40\phcl \\
length & & $-$0.49\phcl \\
span distinct. & & $-$0.22\phcl \\
\multicolumn{2}{l}{boundary distinct.} & 0.16\phcl \\
\bottomrule
\multicolumn{3}{c}{\textbf{Model--span type interactions}}\\
\midrule
\multirow{4}{*}{Handcrafted $\times$} & freq & (0.05)\\
& length & ($-$0.04)\\
& span distinct. & ($-$0.09)\\
& boundary distinct. & (0.09)\\
\midrule
\multirow{4}{*}{CRF $\times$} & freq & $-$0.33\phcl\\
& length & 0.19\phcl\\
& span distinct. & 0.34\phcl\\
& boundary distinct. & $-$0.30\phcl\\
\midrule
\multirow{4}{*}{LSTM $\times$} & freq & 0.47\phcl\\
& length & 0.08\phcl\\
& span distinct. & ($-$0.09)\\
& boundary distinct. & 0.22\phcl\\
\midrule
\multirow{4}{*}{BERT $\times$} & freq & $-$0.43\phcl\\
& length & 0.13\phcl\\
& span distinct. & (0.04)\\
& boundary distinct. & ($-$0.05)\\
\bottomrule
\multicolumn{3}{c}{\textbf{Model--model interactions}}\\
\midrule
\multirow{3}{*}{Handcrafted $\times$} & CRF & 0.10\phcl\\
& LSTM & 0.05\phcl\\
& BERT & $-$0.05\phcl\\
\midrule
\multirow{2}{*}{CRF $\times$} & LSTM & ($-$0.05)\\
& BERT & $-$0.24\phcl\\
\midrule
\multirow{1}{*}{LSTM $\times$} & BERT & $-$0.17\phcl\\
\bottomrule
\end{tabular}
\caption{Regression coefficients from performance prediction model. Coefficients not
  statistically significant at $p < 0.002$ (as per Bonferroni correction) in parentheses.}
\label{tab:coefs}
\end{table}
\paragraph{Frequency helps, length hurts.} The main effects of our
span type predictors show mostly an expected pattern. Frequency has a
strong positive effect (frequent span types are easier to learn),
while length has an even stronger negative effect (long span types are
difficult). More distinct boundaries help performance as well. More
surprising is the negative sign of the span distinctiveness predictor,
which would mean that more distinct spans are more difficult to
recognize. However, this might be due to the negative correlation between
span distinctiveness and frequency ($r=-0.46$ in standardized predictors) --
less frequent spans are, by virtue of their rarity, more distinctive.

\paragraph{BERT is good for performance, especially with few
  examples.} The presence of a BERT component is the highest-impact
positive predictor for model performance, with a positive coefficient
of 1. This finding is not entirely surprising, given the recent
popularity of BERT-based models for span identification problems
\cite{li2020survey,hu-etal-2019-open}. Furthermore, the strong negative value of
the (BERT $\times$ \nspans) predictor shows that BERT's benefits are
strongest when there are few training examples, validating our
hypothesis about transfer learning. BERT is also robust: largely
independent of span or boundary distinctiveness effects.

\paragraph{LSTMs require a lot of data.} While the main effect of
LSTMs is negative, this effect is again modulated by the high positive
coefficient of the (LSTM $\times$ \nspans) interaction. This means
that their performance is highly dependent on the amount of training
data. Also, LSTMs lead to improvements for long span types and those
with distinct boundaries -- properties that LSTMs arguably can pick up
well but that other models struggle with.

\paragraph{CRFs help.} After BERT, the presence of a CRF shows the
second-most positive main effect on model performance. Given the
strong correlation between adjacent tags in a BIO label sequence, it
makes sense that a model capable of enforcing correlations in its
output sequence would perform well. CRFs can also exploit span
distinctiveness well, presumably by the same mechanism.  Surprisingly,
CRFs show reduced effectiveness for highly frequent spans with
distinct boundaries. We believe that this pattern is best considered as
a relative statement: for frequent, well-separated span types CRFs
gain less than other model types.

\paragraph{Handcrafted features do not matter much.} We find neither a
significant main effect of handcrafted features, nor any significant
interactions with span type predictors. Interactions with model
predictors are significant, but rather small. While a detailed
analysis of architecture-wise $F_1$-scores does show that some
architectures, such as pure CRFs, do seem to benefit more from
hand-crafted features (see Table~\ref{tab:full_results} in the Appendix),
this effect diminishes considerably when model components are mixed.

\paragraph{Combining model components shows diminishing returns.} All
interactions between LSTM, CRF, and BERT are negative. This
demonstrates an overlap in these components' utility. Thus, a simple
``maximal'' combination does not always perform best, as
Table~\ref{tab:results} confirms.

\section{Related Work}

Meta-learning and performance prediction are umbrella terms which
comprise a variety of approaches and formalisms in the literature. We
focus on the literature most relevant to our work and discuss the
relationship.

\paragraph{Performance Prediction for Trained Models.}
In NLP, a number of studies investigate predicting the performance of
models that have been trained previously on novel input. An
example is \newcite{chen2009performance} which develops a general
method to predict the performance of a family of language models.
Similar ideas have been applied more recently to machine translation
\citep{ondrej2017findings}, and automatic speech recognition
\cite{elloumi2018asr}, among
others.  While these approaches share our goal of performance
prediction, they predict performance for the same task and model on
new data, while we generalize across tasks and architectures. Thus,
these approaches are better suited to estimating confidence at
prediction time, while our meta-learning approach can predict a model's
performance before it is trained.

\paragraph{AutoML.}
Automated machine learning, or AutoML, aims at automating various
aspects of machine learning model creation, including hyperparameter
selection, architecture search, and feature engineering
\citep{yao2018taking, he2019automl} While the task of performance
prediction does not directly fall within this research area, a model
for predicting performance is directly applicable to architecture
search.  Within AutoML, the auto-sklearn system
\citep{feurer2015efficient} takes an approach rather similar to ours,
wherein they identify meta-features of datasets, and select
appropriate model architectures based on those meta-features.
However, auto-sklearn does not predict absolute performance as we do,
but instead simply selects good candidate architectures via a
$k$-nearest-neighbors approach in meta-feature space.  Other related
approaches in AutoML use Bayesian optimization, including the combined
model selection and hyperparameter optimization of Auto-WEKA
\citep{autoweka} and the neural architecture search of Auto-keras
\citep{jin2019auto}.

\paragraph{Model Interpretability.}
A number of works have investigated how to analyze and explain the
decisions made by machine learning models.  LIME
\citep{mishra2017local} and Anchors \citep{ribeiro2018anchors} are
examples of systems for explaining a model's decisions for specific
training instances.  Other works seek to explain and summarize how
models perform across an entire dataset. This can be achieved
e.g. through comparison of architecture performances, as in
\newcite{nguyen2007comparisons}, or through meta-modeling of trained
models, as was done in \newcite{pmlr-v80-weiss18a}.  Our present work
falls into this category, including both a comparison of architectures
across datasets and a meta-learning task of model performance.

\paragraph{Meta-learning for One- and Few-shot Learning.}
A recent trend is the application of meta-learning to models for one-
or few-shot learning. In this setting, a meta-learning approach is
used to train models on many distinct tasks, such that they can
subsequently be rapidly fine-tuned to a particular task
\cite{finn2017model,santoro2016meta}. While such approaches use the
same meta-learning framework as we do, their task and methodology are
substantially different. They focus on learning with very few training
examples, while we focus on optimizing performance with normally sized
corpora.  Additionally, these models selectively train preselected
model architectures, while we are concerned with comparisons between
architectures.
 
\paragraph{Model and Corpus Comparisons in Survey Papers.}
In a broad sense, our goal of comparison between existing corpora and
modeling approaches is shared with many existing survey
papers. Surveys include quantitative comparisons of existing systems'
performances on common tasks, producing a results matrix very similar
to ours \cite[i.a.]{li2020survey,Yadav2018ASO,Bostan2018}. However,
most of these surveys limit themselves to collecting results across
models and datasets without performing a detailed quantitative analysis
of these results to identify recurring patterns, as we do with
our performance prediction approach.

\section{Conclusion}
In this work, we considered the class of span identification
tasks. This class contains a number of widely used NLP tasks, but no
comprehensive analysis beyond the level of individual tasks is
available. We took a meta-learning perspective, predicting the
performance of various architectures on various span ID tasks in an
unseen-task setup. Using a number of `key metrics' that we developed
to characterize the span ID tasks, a simple linear regression model
was able to do so at a reasonable accuracy. Notably, even though
BERT-based architectures expectedly perform very well, we find that
different variants are optimal for different tasks. We explain such
patterns by interpreting the parameters of the regression model, which
yields insights into how the properties of span ID tasks interact with
properties of neural model architectures. Such patterns can be used
for manual fine-grained model selection, but our meta-learning model
could also be incorporated directly into AutoML systems.

Our current study could be extended in various directions. First, the
 approach could apply the same meta-learning approach to other classes
 of tasks beyond span ID. Second, a larger range of span type metrics
 could presumably improve model fit, albeit at the cost of
 interpretability. Third, we only predict within-corpus performance,
 and corpus-level similarity metrics could be added to make
 predictions about performance in transfer learning.
 
\section*{Acknowledgements}

This work is supported by IBM Research AI through the IBM AI Horizons
Network. We also acknowledge funding from Deutsche
Forschungsgemeinschaft (project PA 1956/4).  We thank Laura Ana Maria
Oberl\"ander and Heike Adel for fruitful discussions.

\appendix

\section{Training Models}
All code used for training span identification and performance prediction models is 
available for download at our project website: \url{https://www.ims.uni-stuttgart.de/data/span-id-meta-learning}.
All text logs generated during training of span identification models are included.

\subsection{Hardware}
All span identification models were trained using GeForce GTX 1080 Ti GPUs.
Training time varied considerably across architectures -- exact training times
for individual experiments are found in the corresponding training logs.

The performance prediction model was trained on a CPU in a few seconds.

\begin{table}[t]
  \centering
  \begin{tabular}{lr}
  \toprule
  \parc & Token POS tag\\
  & Token lemma\\
  & Constituents containing token\\
  & Constituents starting at token\\
  & Constituents ending at token\\
  \midrule
  \riqua & Token POS tag \textdagger\\
  & Token lemma \textdagger\\
  & Is token a quotation mark?\\
  & Is token a quotation mark?\\
  & Is token capitalized?\\
  & Is token all caps?\\
  \midrule
  \conll & Token POS tag \textdagger\\
  \midrule
  \onto & Token POS tag\\
  & Is token capitalized?\\
  & Is token all caps?\\
  & Character bi- and trigrams\\
  & Constituents containing token\\
  & Constituents starting at token\\
  & Constituents ending at token\\
  \midrule
  \chem & Token POS tag \textdagger\\
  & Token lemma \textdagger\\
  & Is token capitalized?\\
  & Is token all caps?\\
  & Is token purely alphabetic?\\
  & Is token all digits?\\
  \bottomrule
  \end{tabular}
  \caption{Hand-crafted features used. Entries marked with a dagger\textdagger{} were predicted using spaCy \citep{spacy2} -- others were either manually annotated, or were exactly specified by the tokens' surface forms}
  \label{tab:features}
\end{table}

\subsection{Tokenization}
For \parc, \onto, and \conll, which include tokenization information,
and we use the datasets' tokenizations directly For \riqua, we use
spaCy \citep{spacy2} to word-tokenize the text.  We found that spaCy's
tokenization performed particularly poorly for \chem, and so for this
corpus we treated all sequences of alphabetic characters as a token,
all sequences of numbers as a token, and all other characters as
single-character tokens.  For \chem, we found that some spans within
the corpus still did not align with token boundaries.  In these cases,
we excluded the spans entirely from the training data, and treated
them as an automatic false-negative for evaluation purposes.

\par For models including a BERT component, tokens were sub-tokenized
using word-piece tokenization \citep{wordpiece} so as to be compatible
with BERT.  The same bag of token features was given to each word
piece.  Models predicted BIO sequences for these sub-tokens, and spans
were only evaluated as correct when their boundaries matched exactly
with the originally-tokenized corpus.

\subsection{Hyperparameters}

\begin{table}
  \centering
\begin{tabularx}{\linewidth}{Xr}
\toprule
Hyperparameter & Value\\
  \cmidrule(r){1-1}\cmidrule(l){2-2}
Input dimensionality & 300 \\
LSTM units & 300 \\
Softmax output layer units & 300\\
CRF units & 300 \\
LSTM layers & 2 \\
LSTM dropout probability & 0.5\\
Learning rate (non-BERT) & $1\times10^{-3}$ \\
Learning rate (BERT) & $2\times10^{-5}$ \\
\bottomrule
\end{tabularx}
\caption{Hyperparameter choices}
\label{tab:hyperparameters}
\end{table}

\par Due to the large number of experiments run, it was infeasible to do a full grid-search for hyperparameters for each architecture-dataset combination.
As such, we tried to pick reasonable values for hyperparameters, motivated by existing literature, prior research, and implementation defaults of existing libraries.
For BERT-based models, our choice of pre-trained model --
`\texttt{bert-base-uncased}' as provided by the HuggingFace Transformers library \cite{Wolf2019HuggingFacesTS} -- fixed some of these hyperparameters for us.
Table \ref{tab:hyperparameters} enumerates the hyperparameter values used for our architectures.

\subsection{Optimizer and Training}
All models were trained with the Adam optimizer \citep{adam}.
For BERT+CRF and BERT+LSTM+CRF, we train the non-BERT parameters as a first training phase, and then fine-tune all parameters jointly as a second training phase.
In these cases, Adam was re-initialized between two training phases.
For training all non-BERT architectures, and for first training phase in the BERT+CRF and BERT+LSTM+CRF architectures, an initial learning rate of 0.001 was used.
For BERT, and for the second training phase in the BERT+CRF and BERT+LSTM+CRF architectures, an initial learning rate of $2\times10^{-5}$ was used.

\subsection{Early Stopping}
To guide early stopping, micro-averaged \fone scores on the development set were computed after every epoch.
These were computed for all span types, including those which were subsequently excluded from our meta-model.
For datasets which had no dedicated development partition, a portion of the training set was held out for this purpose.
After each epoch, model parameters were saved to disk if the development-set \fone score exceeded the best seen so far.
An exponential moving average of these \fone scores was kept, and training terminated when an epoch's \fone score fell below this average.
For BERT+CRF and BERT+LSTM+CRF, this same early stopping procedure was used for both training phases.
The training logs list development set performance at each epoch for each experiment.
\subsection{Features}
Table \ref{tab:features} lists all manual features that were used in models with the ``Feat'' component.

\balance

\section{Full Tables}
\label{sec:full_tables}

Table \ref{tab:full_tasks} lists the span type properties of all span types from all datasets.
Table \ref{tab:full_results} shows average \fone score for each combination of span type and model architecture.\\[\baselineskip]

\balance

\begin{table*}
  
  \centering
  \begin{tabular}{llrrrr}
\toprule
Dataset & Span type & Frequency & Span length & Span dist. & Boundary dist.\\
\midrule
\multirow{7}{*}{\chem} & Abbreviation & 4536 & 1.17 & 3.85 & 0.94 \\
 & Family & 4089 & 1.44 & 3.15 & 0.99 \\
 & Formula & 4445 & 1.98 & 2.50 & 0.99 \\
 & Identifier & 672 & 2.59 & 3.61 & 1.43 \\
 & Multiple & 202 & 6.49 & 2.10 & 1.60 \\
 & Systematic & 6654 & 2.17 & 2.14 & 0.98 \\
 & Trivial & 8832 & 1.15 & 3.64 & 0.86 \\
\midrule
\multirow{7}{*}{\conll} & ADJP & 2060 & 1.22 & 3.13 & 1.22 \\
 & ADVP & 4227 & 1.07 & 3.02 & 0.74 \\
 & NP & 55048 & 1.89 & 0.48 & 0.65 \\
 & PP & 21281 & 1.01 & 2.08 & 0.59 \\
 & PRT & 556 & 1.00 & 4.59 & 2.20 \\
 & SBAR & 2207 & 1.02 & 3.68 & 1.26 \\
 & VP & 21467 & 1.39 & 1.60 & 0.50 \\
\midrule
\multirow{18}{*}{\onto} & Cardinal & 10901 & 1.20 & 3.45 & 0.90 \\
 & Date & 18791 & 1.87 & 2.62 & 0.88 \\
 & Event & 1009 & 2.65 & 3.15 & 1.32 \\
 & Facility & 1158 & 2.33 & 3.54 & 1.22 \\
 & GPE & 21938 & 1.16 & 3.66 & 0.81 \\
 & Language & 355 & 1.03 & 7.26 & 1.99 \\
 & Law & 459 & 2.92 & 3.16 & 1.69 \\
 & Location & 2160 & 1.69 & 4.14 & 1.10 \\
 & Money & 5217 & 2.61 & 3.87 & 1.41 \\
 & NORP & 9341 & 1.04 & 4.85 & 0.98 \\
 & Ordinal & 2195 & 1.00 & 5.99 & 1.39 \\
 & Organization & 24163 & 1.93 & 2.22 & 0.74 \\
 & Percent & 3802 & 2.30 & 4.35 & 1.50 \\
 & Person & 22035 & 1.51 & 3.54 & 1.24 \\
 & Product & 992 & 1.51 & 4.58 & 1.65 \\
 & Quantity & 1240 & 2.25 & 3.79 & 1.35 \\
 & Time & 1703 & 1.95 & 3.50 & 1.24 \\
 & Work of art & 1279 & 2.77 & 2.15 & 1.67 \\
\midrule
\multirow{2}{*}{\parc} & Content & 17416 & 13.86 & 0.15 & 1.73 \\
 & Cue & 15424 & 1.16 & 2.69 & 1.09 \\
\midrule
\multirow{2}{*}{\riqua} & Cue & 2325 & 1.05 & 4.04 & 1.37 \\
 & Quotation & 4843 & 14.06 & 0.22 & 1.67 \\
\bottomrule
\end{tabular}
   \caption{A listing of all span types considered for each dataset, along with their frequency,
geometric mean span length, span distinctiveness, and boundary distinctiveness.
}
\label{tab:full_tasks}
\end{table*}

\begin{table*}
  \centering
  \newcommand{\rt}[1]{\rotatebox{90}{#1}}
\scalebox{0.93}{
  \begin{tabular}{llrrrrrrrrrrrr}
    \toprule
    
& & \rt{Baseline} & \rt{Feat+Baseline} & \rt{CRF}  & \rt{Feat+CRF}  & \rt{LSTM}  & \rt{Feat+LSTM}  & \rt{LSTM+CRF}  & \rt{Feat+LSTM+CRF}  & \rt{BERT}  & \rt{BERT+CRF}  & \rt{BERT+LSTM+CRF}  & \rt{\scalebox{0.8}[1]{BERT+Feat+LSTM+CRF}}  \\
\toprule
\multirow{2}{*}{\rotatebox[origin=c]{90}{\parc}} & Content & 0.0 & 1.6 & 15.5 & 40.0 & 50.3 & 70.4 & 69.1 & 77.2 & 71.7 & 76.6 & 77.8 & \textbf{78.1} \\
 & Cue & 68.0 & 64.9 & 69.1 & 68.5 & 77.1 & 83.3 & 82.2 & 86.3 & 84.4 & 85.8 & \textbf{86.7} & 86.4 \\
\midrule
\multirow{2}{*}{\rotatebox[origin=c]{90}{\scalebox{0.75}{\riqua}}} & Cue & 64.6 & 49.0 & 58.4 & 47.4 & 73.9 & 74.5 & 81.9 & 80.8 & 78.8 & 83.1 & 83.9 & \textbf{84.3} \\
 & Quotation & 0.0 & 0.0 & 5.6 & 6.0 & 76.5 & 90.2 & 89.0 & \textbf{92.3} & 90.3 & 90.6 & 90.0 & 90.2 \\
\midrule
\multirow{7}{*}{\rotatebox[origin=c]{90}{\conll}} & ADJP & 29.1 & 22.8 & 47.0 & 61.9 & 56.6 & 72.8 & 66.3 & 77.2 & 82.4 & \textbf{84.2} & 83.6 & 83.5 \\
 & ADVP & 51.8 & 58.0 & 66.9 & 74.0 & 70.4 & 76.0 & 76.8 & 81.2 & 85.3 & 86.2 & \textbf{86.4} & 86.3 \\
 & NP & 59.5 & 64.3 & 79.7 & 86.0 & 88.8 & 92.8 & 91.4 & 94.9 & 97.0 & 97.1 & 97.2 & \textbf{97.3} \\
 & PP & 57.5 & 56.6 & 90.4 & 94.5 & 94.4 & 96.5 & 96.0 & 97.4 & 98.5 & 98.6 & \textbf{98.6} & 98.6 \\
 & PRT & 40.9 & 41.0 & 64.3 & 63.0 & 66.4 & 68.7 & 73.8 & 75.1 & 84.3 & 83.3 & \textbf{84.6} & 84.6 \\
 & SBAR & 33.2 & 63.3 & 67.1 & 73.8 & 81.4 & 86.1 & 67.1 & 65.1 & 94.2 & 94.3 & 94.2 & \textbf{94.5} \\
 & VP & 49.3 & 56.6 & 74.8 & 89.2 & 84.8 & 92.7 & 88.6 & 94.4 & 96.6 & 96.6 & 96.5 & \textbf{96.7} \\
\midrule
\multirow{18}{*}{\rotatebox[origin=c]{90}{\onto}} & Cardinal & 25.8 & 19.0 & 57.1 & 55.3 & 53.1 & 60.9 & 72.8 & \textbf{81.0} & 80.8 & 80.9 & 79.9 & 79.2 \\
 & Date & 38.6 & 29.0 & 65.6 & 69.0 & 63.1 & 68.3 & 79.5 & 84.3 & 85.5 & \textbf{85.9} & 85.9 & 85.7 \\
 & Event & 0.0 & 0.0 & 29.4 & 40.7 & 0.9 & 0.0 & 39.0 & 46.4 & 63.2 & \textbf{65.0} & 60.5 & 64.9 \\
 & Facility & 0.0 & 0.0 & 7.1 & 17.8 & 0.0 & 0.0 & 30.6 & 45.6 & 62.0 & 64.7 & 64.7 & \textbf{72.8} \\
 & GPE & 60.8 & 44.5 & 75.0 & 76.7 & 69.5 & 72.0 & 85.2 & 91.6 & 93.4 & 94.1 & 94.0 & \textbf{94.7} \\
 & Language & 0.0 & 0.0 & 29.0 & 33.2 & 0.0 & 0.0 & 47.5 & 40.5 & 72.0 & \textbf{76.5} & 71.6 & 70.8 \\
 & Law & 0.0 & 0.0 & 19.1 & 17.9 & 0.0 & 0.0 & 44.7 & 52.1 & 61.5 & 64.8 & 55.7 & \textbf{67.2} \\
 & Location & 11.4 & 0.0 & 38.6 & 42.1 & 19.4 & 9.0 & 54.2 & 67.1 & 65.8 & 67.1 & 66.8 & \textbf{70.8} \\
 & Money & 9.8 & 32.5 & 67.9 & 79.2 & 64.5 & 76.1 & 82.6 & \textbf{90.1} & 87.8 & 88.7 & 89.4 & 88.9 \\
 & NORP & 66.6 & 48.0 & 78.9 & 80.0 & 66.4 & 73.9 & 81.4 & 91.0 & 89.1 & 89.8 & 90.3 & \textbf{91.5} \\
 & Ordinal & 55.6 & 0.4 & 50.9 & 56.0 & 33.1 & 46.8 & 69.3 & \textbf{83.3} & 79.2 & 79.8 & 78.3 & 76.8 \\
 & Organization & 27.3 & 19.1 & 49.1 & 60.6 & 46.3 & 58.6 & 72.5 & 84.0 & 83.8 & 85.4 & 85.6 & \textbf{88.6} \\
 & Percent & 30.1 & 18.8 & 80.0 & 85.8 & 73.8 & 81.3 & 83.3 & \textbf{88.6} & 88.3 & 87.6 & 88.5 & 88.1 \\
 & Person & 25.9 & 15.3 & 52.6 & 71.8 & 64.7 & 70.1 & 79.4 & 87.6 & 92.8 & \textbf{93.7} & 93.3 & 93.6 \\
 & Product & 0.0 & 0.0 & 43.0 & 35.1 & 11.4 & 4.6 & 47.3 & 50.8 & 59.6 & 64.1 & 62.4 & \textbf{64.9} \\
 & Quantity & 0.0 & 0.0 & 38.0 & 49.8 & 25.9 & 0.0 & 64.5 & \textbf{68.0} & 67.0 & 66.7 & 59.2 & 60.6 \\
 & Time & 2.9 & 0.0 & 40.4 & 33.6 & 26.7 & 13.0 & 49.1 & \textbf{63.7} & 61.5 & 61.3 & 60.7 & 61.8 \\
 & Work of art & 0.0 & 0.0 & 6.0 & 7.3 & 0.5 & 17.3 & 29.3 & 57.2 & 55.2 & 59.0 & 57.0 & \textbf{62.4} \\
\midrule
\multirow{7}{*}{\rotatebox[origin=c]{90}{\chem}} & Abbreviation & 50.0 & 12.0 & 54.7 & 54.5 & 51.6 & 48.3 & 62.7 & 71.3 & 78.2 & \textbf{79.1} & 78.2 & 77.1 \\
 & Family & 47.4 & 3.8 & 57.9 & 56.6 & 53.7 & 13.8 & 64.5 & 68.8 & 77.6 & 78.6 & 78.9 & \textbf{79.1} \\
 & Formula & 31.4 & 10.8 & 46.3 & 53.8 & 48.2 & 47.4 & 72.4 & 76.3 & 76.9 & 80.2 & 81.3 & \textbf{81.8} \\
 & Identifier & 0.0 & 0.0 & 44.1 & 37.7 & 38.1 & 0.7 & 69.1 & 66.2 & 79.5 & \textbf{83.1} & 82.5 & 81.8 \\
 & Multiple & 0.0 & 0.0 & 5.1 & 3.0 & 0.0 & 0.0 & 35.5 & 53.4 & 57.8 & 64.6 & 65.6 & \textbf{69.0} \\
 & Systematic & 50.5 & 25.4 & 54.6 & 59.7 & 60.5 & 49.7 & 76.3 & 79.1 & 86.2 & 87.4 & \textbf{87.9} & 87.8 \\
 & Trivial & 61.3 & 30.2 & 66.6 & 65.0 & 65.0 & 52.6 & 75.0 & 77.9 & 90.2 & 91.0 & \textbf{91.2} & 91.1 \\

\bottomrule
\end{tabular}
 }
\caption{\fone scores for each model architecture on each span type.  Each entry is averaged
over five runs.
\label{tab:full_results}
}
\end{table*}

\end{document}